\documentclass[letterpaper, 10 pt, conference]{ieeeconf}  

\IEEEoverridecommandlockouts                              
\title{\LARGE Neural Network-Inspired Analog-to-Digital Conversion to Achieve Super-Resolution with Low-Precision RRAM Devices}

\author{ \normalsize Weidong Cao*, Liu Ke*, Ayan Chakrabarti** and Xuan Zhang*
\\ * Department of ESE, ** Department of CSE, Washington University, St.louis, MO, USA}

\usepackage[bookmarks=false]{hyperref}

\usepackage{cite}
\usepackage{threeparttable}
\usepackage{tabularx}
\usepackage{diagbox}
\usepackage{fancyhdr}
\usepackage[normalem]{ulem}
\usepackage{hyperref}
\usepackage{amsmath}
\usepackage{booktabs} 
\usepackage{upgreek}
\usepackage{graphicx}
\usepackage{verbatim}
\usepackage{multirow}
\usepackage{caption}
\usepackage{array}
\usepackage{subfigure}
\usepackage{float}
\usepackage{CJK}
\usepackage{setspace}
\usepackage[font=footnotesize]{caption}
\usepackage{indentfirst}
\usepackage{dblfloatfix} 
\usepackage{multirow}
\usepackage[noend]{algpseudocode}
\usepackage{algorithmicx,algorithm}
\usepackage{tikz}
\newcommand*\circled[1]
            {\raisebox{.5pt}{\textcircled{\raisebox{-.9pt} {#1}}}}

\graphicspath{{./Fig/}}

  \DeclareSymbolFont{numbers}{T1}{ptm}{m}{n}
  \SetSymbolFont{numbers}{bold}{T1}{ptm}{bx}{n}
  \DeclareMathSymbol{0}\mathalpha{numbers}{"30}
  \DeclareMathSymbol{1}\mathalpha{numbers}{"31}
  \DeclareMathSymbol{2}\mathalpha{numbers}{"32}
  \DeclareMathSymbol{3}\mathalpha{numbers}{"33}
  \DeclareMathSymbol{4}\mathalpha{numbers}{"34}
  \DeclareMathSymbol{5}\mathalpha{numbers}{"35}
  \DeclareMathSymbol{6}\mathalpha{numbers}{"36}
  \DeclareMathSymbol{7}\mathalpha{numbers}{"37}
  \DeclareMathSymbol{8}\mathalpha{numbers}{"38}
  \DeclareMathSymbol{9}\mathalpha{numbers}{"39}

\begin{document}

\maketitle
\pagestyle{empty}
\thispagestyle{empty}
\begin{abstract}

Recent works propose neural network- (NN-) inspired analog-to-digital converters (NNADCs) and demonstrate their great potentials in many emerging applications.
These NNADCs often rely on resistive random-access memory (RRAM) devices to realize the NN operations and require 
high-precision RRAM cells (6$\sim$12-bit) to achieve a moderate quantization resolution (4$\sim$8-bit).
Such optimistic assumption of RRAM resolution, however, is not supported by fabrication data of RRAM arrays in large-scale production process.
In this paper, we propose an NN-inspired super-resolution ADC based on low-precision RRAM devices by taking the advantage of a co-design methodology that combines a pipelined hardware architecture with a custom NN training framework.
Results obtained from SPICE simulations demonstrate that our method leads to robust design of a 14-bit super-resolution ADC using 3-bit RRAM devices with improved power and speed performance and competitive figure-of-merits (FoMs).
In addition to the linear uniform quantization, the proposed ADC can also support configurable high-resolution nonlinear quantization with high conversion speed and low conversion energy, enabling future intelligent analog-to-information interfaces for near-sensor analytics and processing.
\end{abstract}

\section{Introduction}
\label{sec:intro}

Many emerging applications have posed new challenges for the design of conventional analog-to-digital (A$/$D) converters (ADCs)~\cite{c1,c2,c3,c4}.
For example, multi-sensor systems desire programmable nonlinear A$/$D quantization to maximize the extraction of useful features from the raw analog signal, instead of directly performing uniform quantization by conventional ADCs~\cite{c3,c4}.
This can alleviate the computational burden and reduce the power consumption of backend digital processing, which is the dominant bottleneck in intelligent multi-sensor systems. 
However, such flexible and configurable quantization schemes are not readily supported by conventional ADCs with dedicated circuitry that has fixed conversion references and thresholds.

To overcome this inherent limitation of conventional ADCs, several recent works~\cite{c7,c8,c9} have introduced neural network-inspired ADCs (NNADCs) as a novel approach to designing intelligent and flexible A$/$D interfaces.
For instance, a learnable 8-bit NNADC~\cite{c9} is presented to approximate multiple quantization schemes where the NN weight parameters are trained off-line and can be configured by programming the same hardware substrate.
Another example is a 4-bit neuromorphic ADC~\cite{c8} proposed for general-purpose data conversion using on-line training by leveraging the input amplitude statistics and application sensitivity.
These NNADCs are often built on resistive random-access memory (RRAM) crossbar array to realize the basic NN operations, and can be trained to approximate the specific quantization$/$conversion functions required by different systems.
However, a major challenge for designing such NNADCs is the limited conductance$/$resistance resolution of RRAM devices.
Although these NNADCs optimistically assume that each RRAM cell can be precisely programmed with 6$\sim$12-bit resolution, measured data from realistic fabrication process suggest the actual RRAM resolution tends to be much lower (2$\sim$4-bit)~\cite{c10,c11}.
Therefore, there exists a gap between the reality and the assumption of RRAM precision, yet lacks a design methodology to build super-resolution NNADCs from low-precision RRAM devices. 

In this paper, we bridge this gap by introducing an NN-inspired design methodology that constructs super-resolution ADCs with low-precision RRAM devices.
Taking advantage of a co-design methodology that combines a pipelined hardware architecture with deep learning-based custom training framework, our method is able to achieve an NN-inspired ADC whose resolution far exceeds the precision of the underlying RRAM devices.
The key idea of a pipelined architecture is that many consecutive low-resolution (1$\sim$3-bit) quantization stages can be cascaded in a chain structure to obtain higher resolution.
Since each stage now only needs to resolve 1$\sim$3-bit, we can accurately train and instantiate it with low-precision RRAM devices to approximate the ideal quantization functions and residue functions.
Key innovations and contributions in this paper are as follow: 
\vspace{-0.10cm}
\begin{itemize}
	\item We propose a co-design methodology leveraging pipelined hardware architecture and custom training framework to achieve super-resolution analog-to-digital conversion that far exceeds the limited precision of the RRAM device.
	\item We systematically evaluate the impacts of NN size and RRAM precision on the accuracy of NN-inspired sub-ADC and residue block and perform design space exploration to search for optimal pipelined stage configuration with balanced trade-off between speed, area, and power consumption.
    \item SPICE simulation results demonstrate that our proposed method is able to generate robust design of a 14-bit super-resolution NNADC using 3-bit RRAM devices.
    Comparisons with both the state-of-the-art ADCs and other NNADC designs reveal improved performance and competitive figure-of-merits (FoMs).
    \item Our proposed ADC can also support configurable nonlinear quantization with high-resolution, high conversion speed, and low conversion energy.

\end{itemize}

\section{Preliminaries}
\label{sec:bg}

\subsection{RRAM Device, Crossbar Array and NN}
\label{sec:rram}
\subsubsection{RRAM device}A RRAM device is a passive two-terminal element with variable resistance and possesses many special advantages, such as small cell size ($4F^2$, $F$--the minimum feature size), excellent scalability ($<$10$nm$), faster read$/$write time ($<$10$ns$) and better endurance ($\sim$10$^{10}$ cycles) than Flash devices~\cite{c2,c12}. 

\subsubsection{RRAM crossbar array}RRAM devices can be organized into various ultra-dense crossbar array architectures. 
Fig.~\ref{fig:cross}(a) shows a passive crossbar array composed of two sub-arrays to realize bipolar weights without the use of power-hungry operational-amplifiers (op-amps)~\cite{c9}.
The relationship between the input voltage ``vector'' ($\vec{V}_{\text{in}}$) and output voltage ``vector'' ($\vec{V}_{\text{o}}$) can be expressed as $V_{\text{o},j} =\sum\nolimits_{k} W_{k,j} \cdot V_{\text{in},k} + V_{\text{off},j}$.
Here, $k$ ($k\in\{1,2,\ldots,H\}$) and $j$ ($j\in\{1,2,\ldots,M\}$) are the indices of input ports and output ports of the crossbar array.
The weight $W_{k,j}$ can be represented by the subtraction of two conductances in upper ($U$) sub-array and lower ($L$) sub-array as
\vspace{-0.1cm}
\begin{equation}
\label{eq:cross_w}
W_{k,j}=(g^U_{k,j} - g^L_{k,j})/{\sum},~~ \sum=\sum\nolimits_{k}(g^{U}_{k,j}+g^{L}_{k,j}).
\vspace{-0.1cm}
\end{equation}
Therefore, the RRAM crossbar array is capable of performing analog vector-matrix multiplication (VMM) and the parameters of the
matrix rely on the RRAM resistance states.

\subsubsection{Artificial NN}
\begin{figure}
\centering
\includegraphics[width=0.479\textwidth]{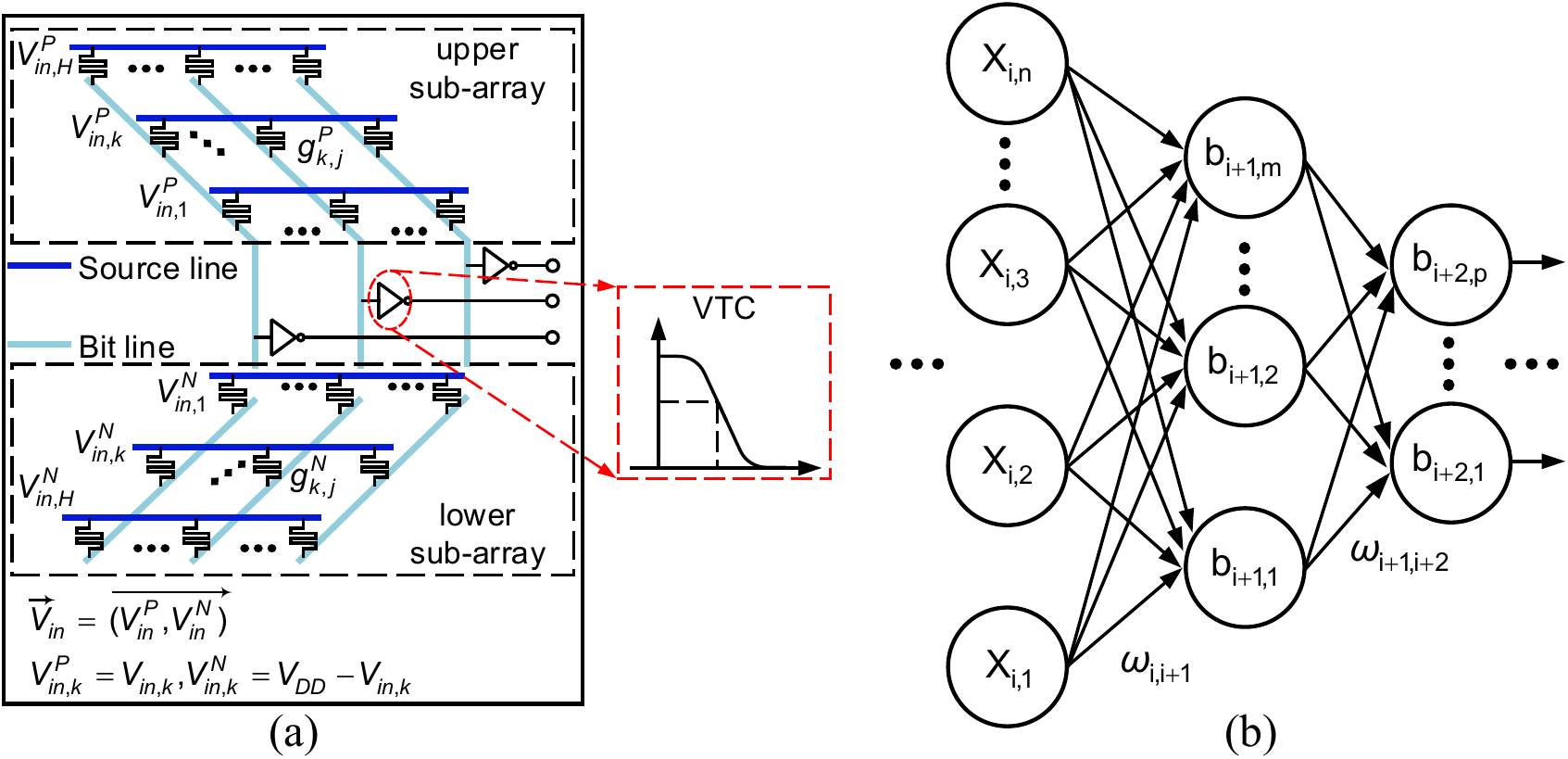}
\vskip -3pt
\caption{(a) Hardware substrate to perform basic NN operations, where the passive crossbar array with two sub-arrays executes VMM and the VTC of CMOS inverter acts as NAF. (b) An example of NN.}
\label{fig:cross}
\vskip -15pt
\end{figure}

With the RRAM crossbar array, an NN shown in Fig.~\ref{fig:cross}(b) can be implemented on such hardware substrate.
Generally, the NN processes the data by executing the following operations layer-wise~\cite{c23}:
\vspace{-0.1cm}
\begin{equation}
\label{eq:nn_c}
\vec{y}_{i+1} =f( W_{i,i+1} \cdot \vec{x}_i + \vec{b}_{i+1}).
\vspace{-0.1cm}
\end{equation}
Here, $W_{i,i+1}$ is the weight matrix to connect the layer $i$ and layer $(i+1)$.
$f(\cdot)$ is a nonlinear activation function (NAF).
These basic NN operations, e.g., VMM and NAF, can be mapped to the RRAM crossbar array and CMOS inverters shown in Fig.~\ref{fig:cross}(a), where the voltage transfer characteristic (VTC) is used as an NAF~\cite{c9}.

\subsection{NN-Inspired ADCs}
\label{sec:nnadc}

ADC can be viewed as a special case of classification problems which maps a continuous analog signal to a multi-bit digital code.
An NN can be trained to learn this input-output relationship, and a hardware implementation of this NN can be instantiated in the analog and mixed-signal domain.
This is the basic idea behind NNADCs which implements the learned NN on a hardware substrate to approximate the desired quantization functions for data conversion:
\vspace{-0.1cm}
\begin{equation}
\label{eq:adc}
\sum\limits_{i=1}^{M} 2^{i-1} \cdot b_{\text{o},i} = \text{round}\left( \frac{V_{\text{in}} - V_{\min}}{V_{\max}-V_{\min}} \times (2^M-1)  \right),
\vspace{-0.1cm}
\end{equation}
where, $M$ is the resolution; $V_{\text{in}}$ is input analog signal and $b_{\text{o}}$ is the output digital codes; $V_{\min}$ and $V_{\max}$ are the minimum and maximum values of the scalar input signal $V_{\text{in}}$.
Since RRAM crossbar array provides a promising hardware substrate to build NNs, recent work has demonstrated several NNADCs based on RRAM devices~\cite{c7,c8,c9}.
Although the NN architectures adopted by these NNADCs are various, they all rely on a training process to learn the appropriate NN weights to approximate flexible quantization schemes that can be configured by programming the weights stored in RRAM conductance$/$resistance.
However, existing NNADCs~\cite{c7,c8,c9} often exhibit modest conversion resolution (4$\sim$8-bit) and invariably rely on optimistic assumption of the RRAM precision (6$\sim$12-bit), which is not well substantiated by measurement data from realistic RRAM fabrication process~\cite{c10,c11}.
This resolution limitation severely constrains the application of NNADCs in emerging multi-sensor systems that require high-resolution ($>$10-bit) A$/$D interfaces for feature extraction and near-sensor processing~\cite{c1,c3,c4}.

\subsection{Pipelined ADCs}
\label{sec:pipeadc}

\begin{figure}
\centering
\includegraphics[width=0.479\textwidth]{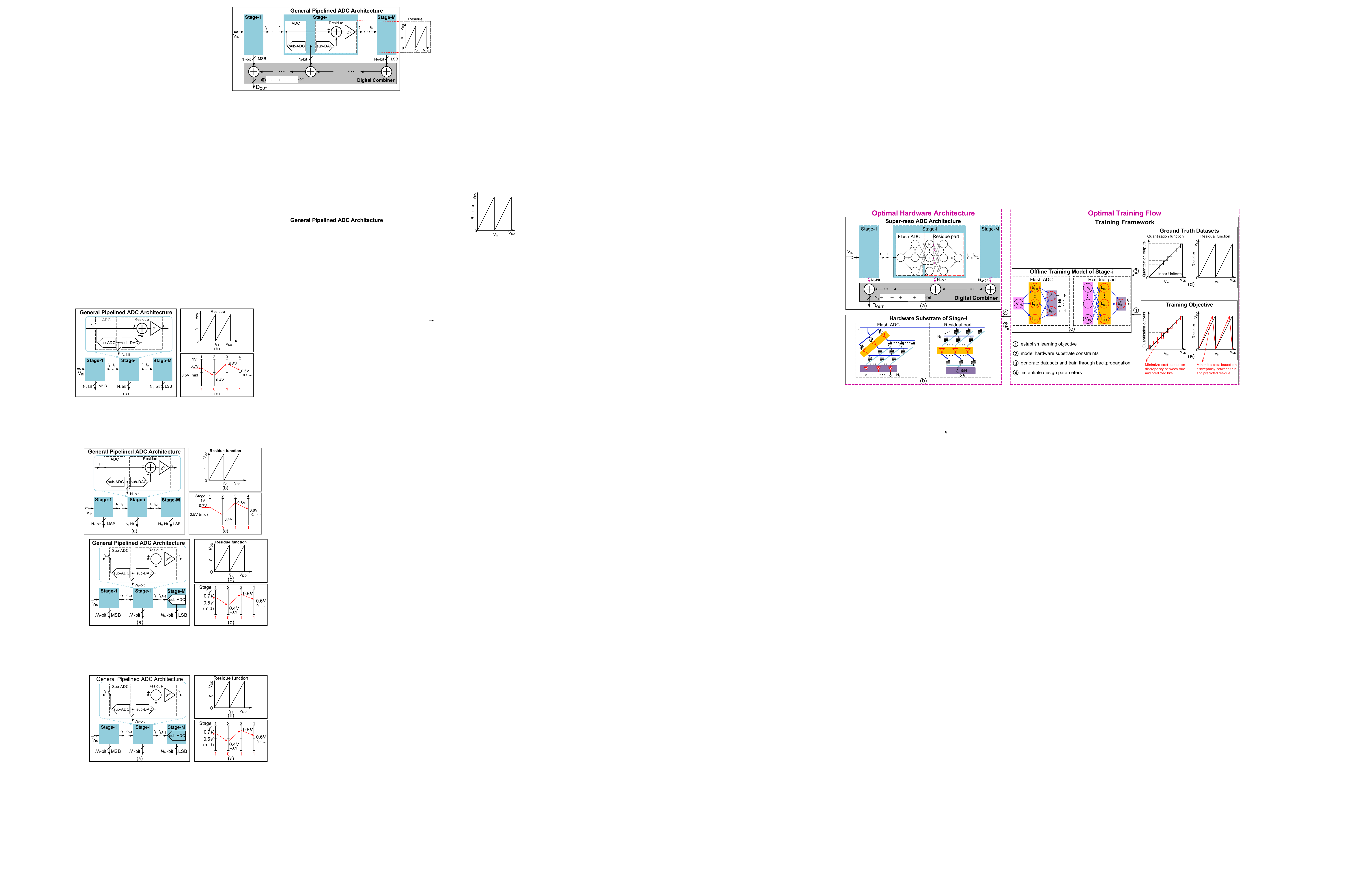}
\vskip -3pt
\caption{(a) General architecture of pipelined ADC. (b) An example of residue function when $N_i=1$. (c) A quantization example of a 4-bit pipelined ADC with four 1-bit stages.}
\label{fig:pipeadc}
\vskip -15pt
\end{figure}

Pipelined architecture is a well-established ADC topology to achieve high sampling rate and high resolution with low-resolution quantization stages~\cite{c15}.
Fig.~\ref{fig:pipeadc}(a) illustrates a typical pipelined ADC with $M$ stages whose resolution $\text{RESO}$ can be achieved by concatenating $N_i$-bit of each stage with digital combiner: $\text{RESO}=\sum^{M}_{i=1}N_i$.
Note that $N_i$ is usually $\leq4$ and not necessarily identical in all stages.
As the Fig.~\ref{fig:pipeadc}(a) illustrates, an arbitrary stage-$i$ contains two sub-blocks: a sub-ADC and a residue.
The sub-ADC resolves $N_i$-bit binary codes $D_{N_i}$ from input residue $r_{i-1}$, while the residue part amplifies the subtraction between the input residue $r_{i-1}$ and the analog output of sub-ADC by $2^{N_i}$ to generate the output residue $r_i$ for next stage.
This process can be expressed as a simple function:
\vspace{-0.1cm}
\begin{equation}
\label{eq:re}
r_i = [r_{i-1}-\textit{REF}(D_{N_i})]\cdot 2^{N_i}.
\vspace{-0.1cm}
\end{equation}
Here, $\textit{REF}(D_{N_i})$ is the analog output of sub-DAC that depends on $D_{N_i}$. 
For example, assuming $r_{i-1}\in[0,V_{\text{DD}}]$ and $N_i=1$, then $\textit{REF}(0)=0$ and $\textit{REF}(1)=V_{\text{DD}}/2$;
and Fig.~\ref{fig:pipeadc}(b) shows the corresponding residue function.
To understand the basic working principle of pipelined ADCs, we use a 4-bit pipelined ADC with four 1-bit stages in Fig.~\ref{fig:pipeadc}(c) as an example. 
Assuming the initial analog input is $0.7V$ ($V_{\text{DD}}=1V$), then the first stage will output ``1''---a digital code, and ``$0.4V$''--- an analog residue according to Eq.~\eqref{eq:re} which will be processed by the following stage in the same way as initial analog input.
Finally, we can obtain 4-bit outputs $1011$, which is the quantization of $0.7V$ ($0.7/1=11.2/2^4\approx11/2^4$).
This example also shows that a higher resolution (4-bit) can indeed be constructed with low-precision (1-bit) stages in a pipelined ADC.

\section{Co-Design Methodology}
\label{sec:DM}
\newcommand{\sigvtc}{\sigma_{\text{\tiny VTC,i}}}
\newcommand{\vcmp}{V_{\text{\tiny cmp}}}
\newcommand{\bgt}{r_{\text{\tiny GT}}}
\renewcommand{\Re}{\mathbb{R}}
\label{sec:training}



\begin{figure*}[htbp]
\centering
\includegraphics[width=1\textwidth]{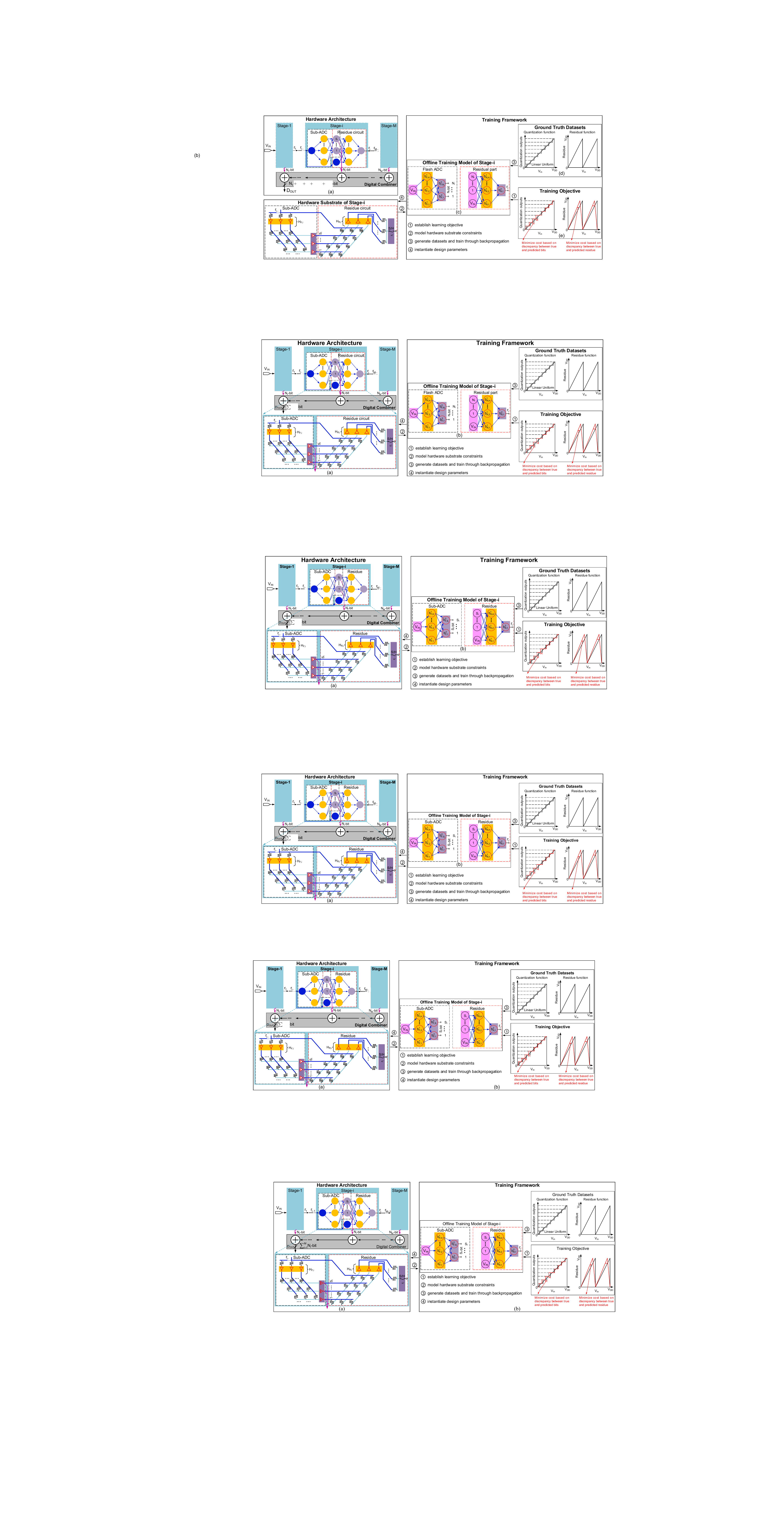}
\vskip -3pt
\caption{Proposed co-design framework for the super-resolution NNADC. (a) Pipelined architecture for the proposed NNADC. (b) Off-line training model of each stage-$i$. Proposed training framework takes ground truth datasets as inputs during off-line training to find the optimal weights and derive the RRAM resistances to minimize cost function and best approximate ideal quantization function and residue function.}
\label{fig:designflow}
\vskip -12pt
\end{figure*}

\subsection{Hardware Substrate}
\label{sec:HSF}

\subsubsection{Pipelined architecture}The observation from traditional pipelined ADCs motivates us to extend such architecture to NNADC to enhance its resolution beyond the limit of RRAM precision.
The overall hardware architecture for the proposed high-resolution NNADC is presented in Fig.~\ref{fig:designflow}(a), where a pipelined architecture composed of cascaded conversion stages is adopted in the design.
This pipelined architecture brings two direct benefits.
First, each stage in the proposed NNADC now only needs to resolve $1$$\sim$$3$-bit quantization, which is well within the precision limit of current RRAM fabrication process~\cite{c10,c11} and can be easily achieved with the automated design methodology introduced in previous work~\cite{c9}.
Second, although many cascading stages are needed, there only exist three distinct low-resolution configurations to choose from for each stage, namely $N_i=1,2,3$.
This allows us to simplify the design process by focusing on optimizing the sub-block design of each stage with different resolutions.
The full pipelined system can then be assembled by iterating through different combinations of the sub-blocks with different resolutions.

\subsubsection{Low-resolution NNADC stage}For stage-$i$ in the proposed NNADC, we use a five-layer NN to implement the sub-ADC and the residue block.
The five-layer NN can be decomposed into two three-layer sub-blocks, and each of them can be mapped into the corresponding sub-ADC and residue in Fig.~\ref{fig:pipeadc}(a).
The cornerstone of this mapping methodology is the universal approximation theorem that a feed-forward three-layer NN with a single hidden layer can approximate arbitrary complex functions~\cite{c17}.
We use the RRAM crossbar array and CMOS inverter illustrated in Fig.~\ref{fig:cross}(a) as the hardware substrate to design the sub-blocks of each stage.
As Fig.~\ref{fig:designflow}(b) shows, for the sub-ADC, the input analog signal represents the single ``place holder'' neuron in MLP's input layer.
Therefore, the weight matrix dimensions are $H_{F,i}\times 1$ between the hidden and the input layer, and $H_{F,i}\times S_{i}$ between the hidden and the output layer, assuming there are $H_{F,i}$ and $S_i$ neurons in the hidden and output layer.
Here, we use a redundant ``smooth'' $S_i\rightarrow N_i$ encoding method to replace the standard $N_i$-bit binary encoding with $S_i$ bits ($S_i > N_i$) according to previous work~\cite{c9}, as it improves the training accuracy and reduces hidden layer size of the sub-ADC.
For example, we use $3\rightarrow 2$ smooth codes to train a 2-bit sub-ADC with 3-bit smooth codes as output in Fig.~\ref{fig:residue}(b).
For the residue, there are ($1+S_i$) input neurons (one analog input and $S_{i}$-bit smooth digital codes from the proceeding sub-ADC block), and only one analog output neuron;
therefore, the weight matrix dimensions are $H_{R,i}\times(1+S_i)$ between the hidden and the input layer and $H_{R,i}\times 1$ between the hidden and the output layer, assuming there are $H_{R,i}$ hidden neurons.
The sampling$/$hold (S$/$H) circuits~\cite{c25} are used in the output layer to drive the next stage.
Since the op-amps in Fig.~\ref{fig:pipeadc}(a) are eliminated in the NN-inspired design of residue circuit, considerable power saving can be obtained from each stage.

\subsection{Training Framework}
\label{sec:AF}

\subsubsection{Training overview}We propose a training framework that accurately captures the circuit-level behavior of the hardware substrate in its mathematical model and is able to learn the robust NNs and its associated hardware design parameters (i.e., RRAM conductance) to approximate the sub-ADC and residue for each stage.
The training framework incorporates two important features.
First, we employ collaborative training for the two sub-blocks in each stage. 
The sub-ADC is initially trained to approximate the ideal quantization function with high-fidelity, then its digital outputs and original analog input are directly fed to the residue block for the residue training.
This collaborative training flow can effectively minimize the discrepancy between the circuit artifacts and the ideal conversion at each stage.
Second, non-idealities of devices, such as process, voltage and temperature (PVT) variations of the CMOS device and limited precision of the RRAM devices, can be incorporated into training to make the proposed NNADC robust to these defects~\cite{c18}.
This is another advantage of the proposed NNADC over traditional ADC designs, where even with delicate calibration techniques, the non-idealities cannot be fully mitigated~\cite{c15}.

\subsubsection{Training steps}The detailed training flow is shown in Fig.~\ref{fig:designflow}(b), which consists of four steps.
We focus on describing the training steps for the residue block, as we adopt similar sub-ADC training method that has been elaborated in previous work~\cite{c9,c18}.

Step \circled{1}: establish learning objective. 
For the residue circuit, its output is an analog value; 
therefore, the hardware substrate can be modeled as a three-layer NN with a ``place-holder'' output neuron:
\vspace{-0.1cm}
\begin{equation}
\label{eq:di}
 \tilde{h}_i = L_1(r_{i-1}, D_{S_i};\theta_{1,i}), ~~r_{i} = L_2(h_i;\theta_{2,i}).
\vspace{-0.1cm}
\end{equation}
Here, $h_i = \sigvtc(\tilde{h}_i)$.
$D_{S_i}$ indicates the digital output of the ADC (``1'' means $V_{\text{DD}}$, and ``0'' means $\text{GND}$), and $r_{i-1}$ is the scalar residue input of stage-$i$;
$\tilde{h}_i$ denote the outputs of the first crossbar layer, which are modeled as a linear function $L_1$ of $r_{i-1}$ and $D_{S_i}$, with learnable parameters $\theta_{1,i}=\{W_1,V_1\}$ corresponding to RRAM crossbar array conductances.
Each of these voltages is passed through an inverter (shown in Fig.~\ref{fig:cross}(a)), whose input-output relationship is modeled by the nonlinear function $\sigvtc(\cdot)$, to yield the vector $h_i$.
The linear function $L_{2}$ models the second layer of the crossbar to produce the output residue $r_i$ for next stage, with learnable parameters $\theta_{2,i}=\{W_2,V_2\}$.
The learning objective is to find optimal values for the parameters $\{\theta_{1,i},\theta_{2,i}\}$ such that for all values of $r_{i-1}$ in the input range, the circuit yields corresponding residue $r_{i}$ that are equal or close to the desired ``ground truth'' $\bgt$ in Eq.~\eqref{eq:re}.
To achieve this aim, we define a cost function $C(r_{i},\bgt)$ to measure the discrepancy between predicted $r_{i}$ and true $\bgt$ based on the mean-square loss:
\vspace{-0.1cm}
\begin{equation}
  \label{eq:xent}
  C(r_{i},\bgt) = \sum\nolimits_{j} [{\bgt(j)}-r_i(j)]^2.
\vspace{-0.2cm}
\end{equation}

Step \circled{2}: model hardware constraints.
Hardware constraints come from three aspects: CMOS neuron PVT variations, limited precision of RRAM device, and passive crossbar array.
To reflect these hardware constraints, we first group all VTCs obtained by Monte Carlo simulations in $A_{\text{VTC}}$ using the technology specification in Section~\ref{sec:set}.
Meanwhile, we control the precision of weight with $A_\textit{R}$-bit during the training.
Finally, we let the summation of all elements (absolute value) in each column (``0'') of $W_{1,2}$ be $<$ 1:
\vspace{-0.1cm}
\begin{equation}
\label{eq:wt}
  \sum(abs(W_1),0)<1;~~  \sum(abs(W_2),0)<1,
\vspace{-0.1cm}
\end{equation}
to reflect the weights constraints in Eq.~\eqref{eq:cross_w}.


Step \circled{3}: hardware-oriented training.
We initialize the parameters $\{\theta_{1,i}, \theta_{2,i}\}$ randomly, and update them iteratively based on gradients computed on mini-batches of $\{(r_{i-1},D_{N_i},\bgt)\}$ pairs randomly sampled from the input range.
To incorporate the hardware constraints in step \circled{2} into training, we let each neuron $j$ in Eq.~\eqref{eq:di} randomly pick up a VTC from $A_{\text{VTC}}$ during training:
\vspace{-0.1cm}
\begin{equation}
\label{eq:randvtc}
\sigvtc^j = A_{\text{VTC}}[\text{randint}(N)], j=1,2,...,H_{R,i}.
\vspace{-0.1cm}
\end{equation}
We then periodically clip all values of $W_1$ between $[-1/(1+N_i),1/(1+N_i)]$, as well as $W_2$ between $[-1/H_{R,i},1/H_{R,i}]$ to satisfy Eq.~\eqref{eq:wt}.

Step \circled{4}: instantiate conductance values.
We adopt the same instantiation method based on previous work~\cite{c9}, which is proven to always find a set of equivalent conductances from the trained weights and biases to map to the RRAM devices in the hardware substrate.
After this, we perturb each resistance $\textit{R}$ by:
\vspace{-0.1cm}
\begin{equation}
\label{eq:wup}
\textit{R} \leftarrow \textit{R}\cdot e^{\theta};~~\theta \sim N(0,\sigma^2),
\vspace{-0.1cm}
\end{equation}
to evaluate the robustness of the NN model to the stochastic variation of RRAM resistance~\cite{c2}.

\begin{figure}
\centering
\includegraphics[width=0.479\textwidth]{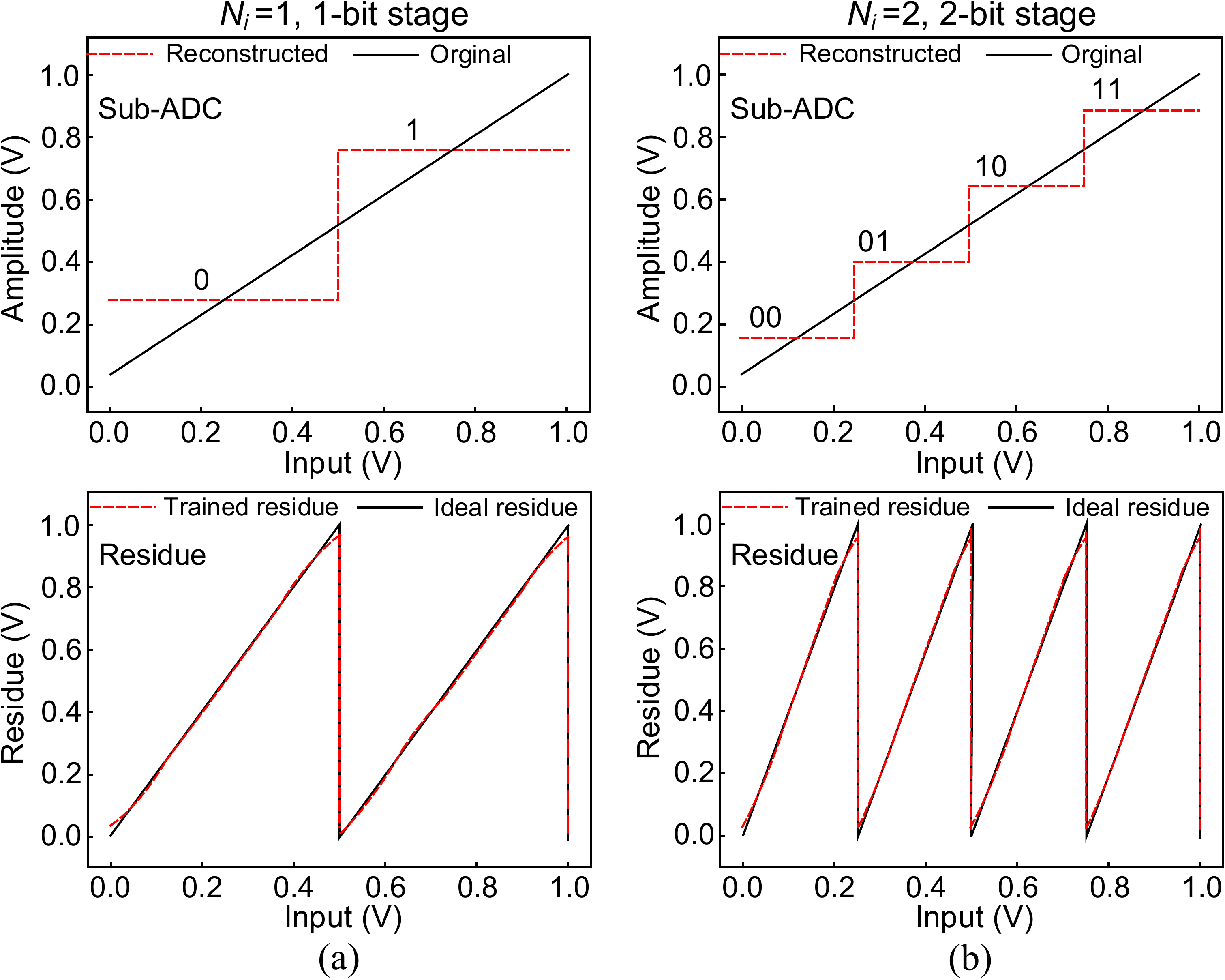}
\vskip -3pt
\caption{Illustrations of trained sub-ADC and residue functions  for a pipeline stage with different resolution. (a) 1-bit stage ($N_i=1$). (b) 2-bit stage ($N_i=2$).}
\label{fig:residue}
\vskip -15pt
\end{figure}

\subsection{Examples of Trained Sub-ADC and Residue}
\label{sec:training_example}
Fig.~\ref{fig:residue} illustrates the SPICE simulation of different trained stages with the proposed training framework.
The sub-ADC and the residue in Fig.~\ref{fig:residue}(a) are trained through a $1\times3\times2$ NN and a $3\times5\times1$ NN respectively by setting $N_i=1$, while the sub-ADC and the residue in Fig.~\ref{fig:residue}(b) are trained through a $1\times4\times3$ NN and a $4\times7\times1$ NN by setting $N_i=2$.
In both figures, we use 3-bit RRAM and set $\sigma=0.05$ in Eq.~\eqref{eq:wup} for evaluation.
The comparison between the trained function and the ideal function shows that each stage with low-precision RRAM can accurately approximate the ideal stage function with the aid of the proposed training framework. 



\section{Experimental Results}
\label{sec:results}

\subsection{Experimental Methodology}
\label{sec:set}

\subsubsection{Training configuration}We set $N_i=1,2,3$ to get three distinct resolution configurations in each pipeline stage in our experiments.
For each stage, we train different NN models and each NN model is trained via stochastic gradient descent with the Adam optimizer using TensorFlow~\cite{c20}.
The weight precision $A_\textit{R}$ during training is set to be 1$\sim$7-bit.
The batch size is 4096, and the projection step is performed every 256 iterations.
We train for a total of 2$\times10^4$ iterations for each sub-ADC model and residue model, varying the learning rate from $10^{-3}$ to $10^{-4}$ across the iterations.

\subsubsection{Technology model}We use the HfO$_x$-based RRAM device model to simulate the crossbar array~\cite{c22}.
We set the resistance stochastic variation $\sigma=0.05$, since it is a moderate variation based on the evaluations from prior work~\cite{c23}.
The transistor model is based on a standard 130$nm$ CMOS technology.
The inverters, output comparators, and transistor switches in the RRAM crossbars are simulated with the 130nm model using Cadence Spectre.
The VTC group $A_{\text{VTC}}$ is obtained by running 100 times Monte Carlo simulations.
The simulation results presented in the following section are all based on SPICE simulation.

\subsubsection{Metric of training accuracy}The trained accuracy of the sub-ADC$/$proposed NNADC is represented by the effective number of bits (ENOB)--a metric to evaluate the effective resolution of an ADC.
We report ENOB based on its standard definition ENOB$=$(SNDR-1.76)$/$6.02, where the signal to noise and distortion ratio (SNDR) is measured from the sub-ADC's$/$proposed NNADC's output spectrum.
The training accuracy of the residue circuit is represented by the mean-square error (MSE) between predicted residue function and ideal residue function.
We report the MSE based on 2048 uniform sampling points in the full range of input $[0,V_{\text{DD}}]$.


\begin{figure}
\centering
\includegraphics[width=0.479\textwidth]{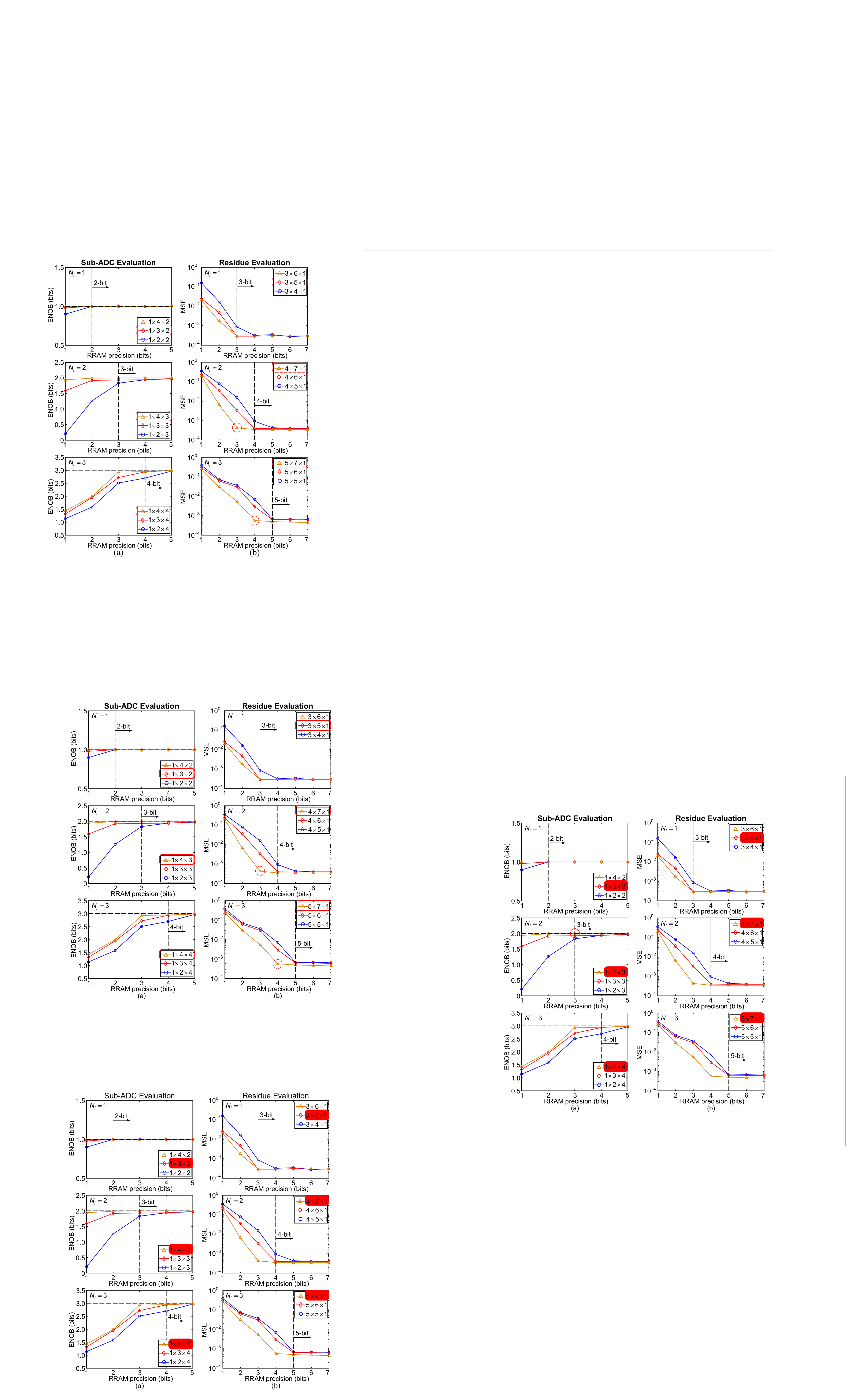}
\vskip -3pt
\caption{Sub-block training performance using different NN models and RRAM precision at a fixed stochastic variation $\sigma=0.05$. (a) The trend between ENOB and RRAM precision of sub-ADC under different NN models, where the $N_i$ is set as 1, 2, 3 respectively. (b) The trend between MSE and RRAM precision of residue circuit under different NN models, where the $N_i$ is set as 1, 2, 3 respectively.
}
\label{fig:rrtrend}
\vskip -15pt
\end{figure}

\subsection{Sub-block Evaluations}
\label{sec:designexp}

\subsubsection{Resolution and robustness}To find a robust design for each stage, we study the relationship between the trained accuracy and RRAM precision of each sub-block with different NN sizes at a fixed stochastic variation.
For these experiments, we first incorporate both CMOS PVT variations and limited precision of RRAM device into training, and then instantiate several batches of 100-run Monte Carlo simulations with a resistance variation $\sigma=0.05$ in Eq.~\eqref{eq:wup}, and finally compute the median accuracy of each model.

We plot the trends in Fig.~\ref{fig:rrtrend}.
Generally, an $(N_i+1)$-bit RRAM precision is enough to train an NN model to accurately approximate an $N_i$-bit sub-ADC, which confirms the conclusion in previous work~\cite{c9}.
Particularly, larger size NN models with more hidden neurons can even accurately approximate an $N_i$-bit sub-ADC with $N_i$-bit RRAM precision.
Similar conclusions can also be made from the trained performance of residue circuits.
As the Fig.~\ref{fig:rrtrend}(b) shows, an $(N_i+2)$-bit RRAM precision is enough to train an NN model to accurately approximate a residue circuit.
Moreover, a larger size NN with more hidden layer neurons can accurately approximate the residue circuit of $N_i$-bit stage with $(N_i+1)$-bit RRAM precision.

\subsubsection{Sub-block design trade-off}Each stage-$i$ has design trade-off among power consumption $P_i$, sampling rate $f_{S,i}$ and area $A_{s,i}$.
A completed design space exploration may involve the searching of different NN sizes of each sub-block in stage-$i$, RRAM precision and stochastic variations.
Here, we use three pairs of sub-blocks highlighted by the solid boxes in Fig.~\ref{fig:rrtrend} as an example to illustrate the design trade-off, since each of them shows enough accuracy and robustness with no more than 4-bit RRAM precision.
For these experiments, we combine each pair of sub-blocks to form three distinct sub-blocks with resolution $N_i= 1,2,3$, respectively.
We then fix the precision of RRAM device with 3-bit for for all building blocks except for the residue in $N_i=3$ stage, which use 4-bit RRAM device.
We finally study the relationship between the power $E_j$, speed $f_{j}$, and area $A_j$ of each distinct stage-$j$ ($j=1,2,3$) by simulating the minimum power consumption/area of each distinct stage that works well at different sampling rates.

\begin{figure}
\centering
\includegraphics[width=0.475\textwidth]{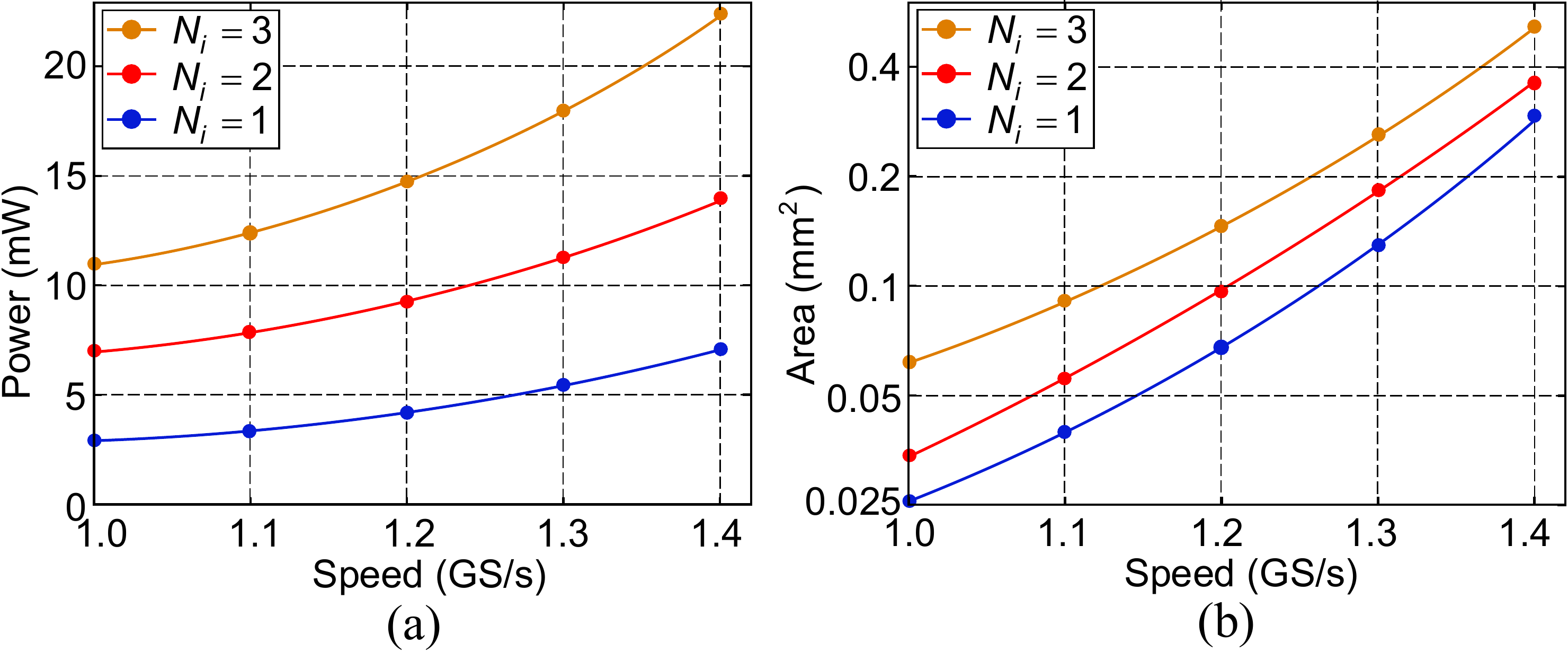}
\vskip -4.9pt
\caption{Design trade-offs of three distinct stages, with resolution $N_i=1,2,3$ respectively. (a) Power VS speed. (c) Area VS speed.}
\label{fig:trade}
\vskip -15pt
\end{figure}

The trends are plotted in Fig.~\ref{fig:trade}, which shows clear trade-offs between speed and power consumption, as well as speed and area, for each distinct stage.
This is because in order to make each sub-block work well under faster speed, we need to increase the driving strength of the neurons by sizing up the inverters, which results in an increase of power consumption and area for each stage.

\subsubsection{Design optimization}Based on the exploration of different sub-block configurations, an optimal design for the proposed ADC with a given resolution can be derived by solving the following optimization problem:
\vspace{-0.10cm}
\begin{equation}
\label{eq:opt}
\begin{split}
&\min\quad FoM_W = P/({2^{ENOB}\cdot f_S})\\
&\min\quad A_{\textit{ADC}}\\
&s.t.\quad  \left\{\begin{array}{lc}
ENOB \leq \sum\limits_{i=1}^M N_i\ & N_i\in\{1,2,3\}, \\
P = \sum\limits_{i=1}^M P_i & P_i\in\{E_1,E_2,E_3\}, \\
f_S = \min\limits_{1\leq i\leq M} \{f_{S,i}\} & f_{S,i}\in \{f_{1},f_{2},f_{3}\}, \\
A_{\textit{ADC}} = \sum\limits_{i=1}^M A_{s,i} & A_{s,i}\in\{A_1,A_2,A_3\}. \end{array}\right.
\end{split}
\vspace{-0.10cm}
\end{equation}
Here, the first optimal objective ${FoM_W}$ ($\textit{fJ}/\textit{conv}$) is a standard figure-of-merit that describes the energy consumption of one conversion for an ADC, and the second optimal objective $A_{\textit{ADC}}$ is the area of the proposed ADC.
We set $FoM_W$ as the main optimal objective, since energy efficiency usually is the most important consideration for most applications.
In this way, as shown in Fig.~\ref{fig:pipe}, we can obtain an optimal design for a maximum 14-bit pipelined NNADC with 12.5 bits of ENOB, and $11.6\textit{fJ}/\textit{conv}$ of $FoM_W$ working at 1$\textit{GS}/\textit{s}$.
It showcases the advantages of our proposed co-design framework that incorporates many circuit-level non-idealities in the training process, allowing us to realize a robust design cascading up to eleven stages, a level often unattainable with traditional pipelined ADCs.

\subsection{Full Pipelined NNADC Evaluation}
\label{sec:fea}

\begin{figure}
\centering
\includegraphics[width=0.479\textwidth]{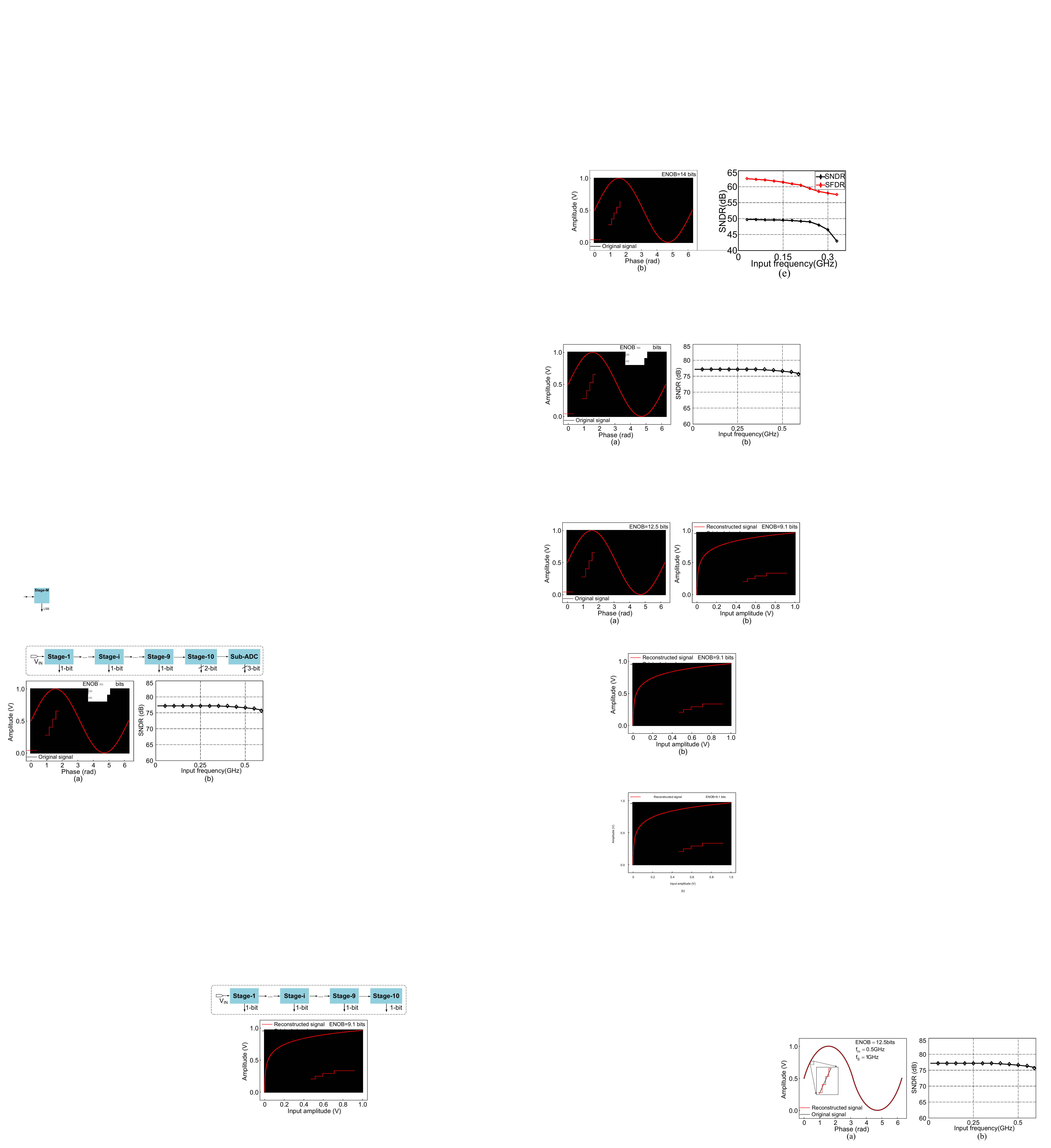}
\vskip -3pt
\caption{(a) Reconstruction of a 14-bit pipelined NNADC with 3-bit RRAM whose pipelined chain consists of eleven stages: nine 1-bit stages, one 2-bit stage and one 3-bit sub-ADC. (b) SNDR trend.}
\label{fig:pipe}
\vskip -6pt
\end{figure}

We choose the three distinct stages in Section~\ref{sec:designexp} to evaluate the quantization ability of the proposed full pipelined NNADC.
We find that although the co-design framework can help us to train a low-resolution stage to approximate the ideal quantization function and residue function with high-fidelity, the minor discrepancy between the trained stage and ideal stage will propagate and aggregate along the pipeline and finally results in a wrong quantization.
Our simulations based on various combinations of different pipeline stages show that a maximum 14-bit pipelined NNADC working at 1$\textit{GS}/\textit{s}$ can be achieved by cascading nine 1-bit stages, one 2-bit stage and one 3-bit sub-ADC with 3-bit RRAM precision.
Note that the last stage of the 14-bit pipelined NNADC does not need to generate residue.
The reconstructed signal of this 14-bit ADC is shown in Fig.~\ref{fig:pipe}(a), where the ENOB is 12.5 bits under $1\textit{GH}$z sampling frequency.
We also report the SNDR trend with input signal frequency in Fig.~\ref{fig:pipe}(b).
The SNDR begins to degenerate after $0.5\textit{GH}$z input, verifying the sampling frequency ($\times2$ of input signal frequency) of the proposed 14-bit NNADC is well above 1$\textit{GH}$z.

\begin{figure}
\centering
\includegraphics[width=0.239\textwidth]{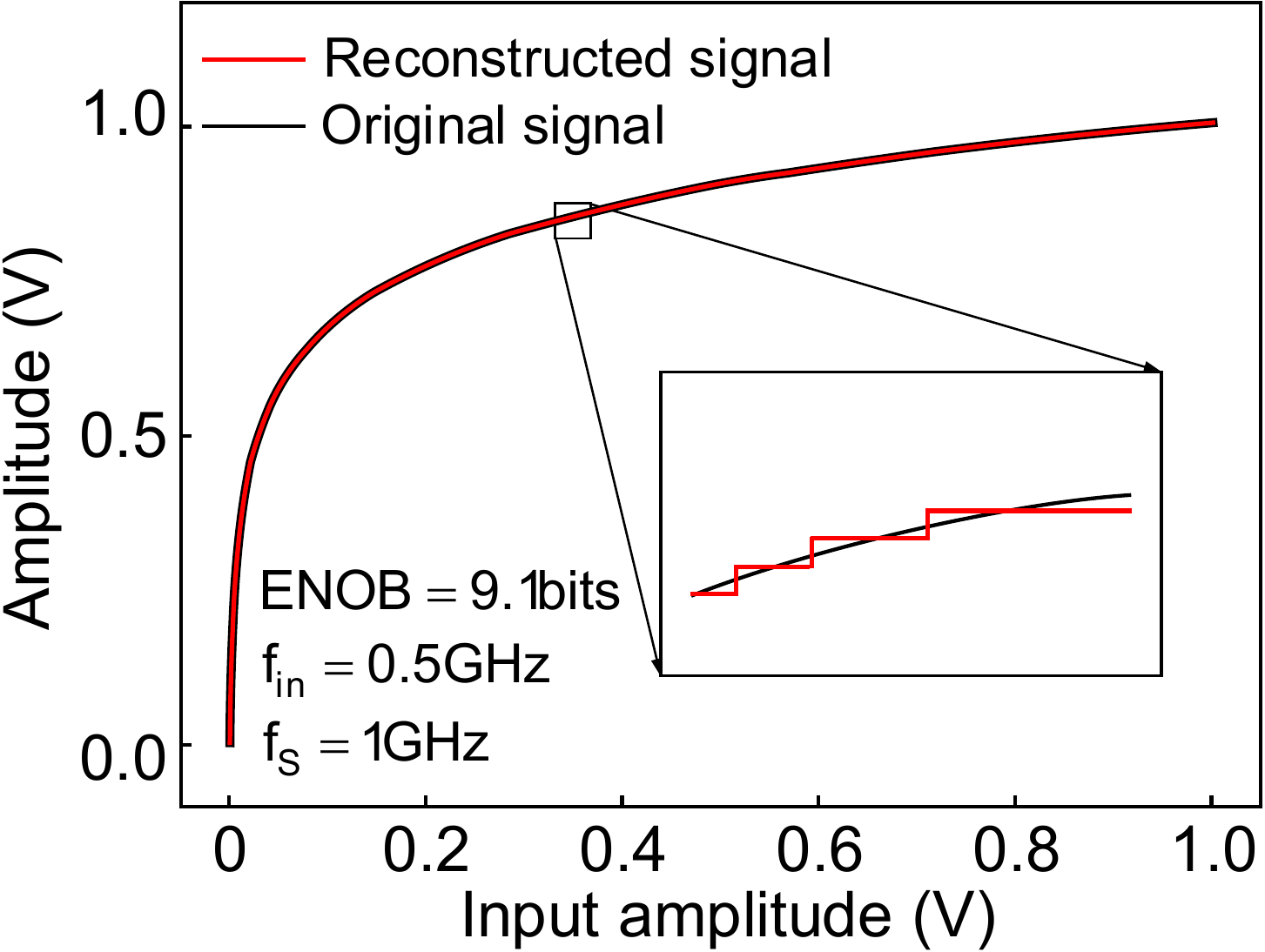}
\vskip -3pt
\caption{A 10-bit logarithmic NNADC with ten 1-bit stages.}
\label{fig:nonlinear}
\vskip -15pt
\end{figure}

Finally, we train a nonlinear ADC based on the same methodology using a logarithmic encoding on the input signal by replacing $V_{\text{in}}$ in Eq.~\eqref{eq:adc} with $V_{\text{in},\log} = V_{\text{DD}}\cdot \log_2(a +1)$ ($a\in\{0,1\}$) to train a 1-bit stage.
We find that a 10-bit logarithmic ADC with 9.1-bit ENOB working at $1\textit{GS}/\textit{s}$ can be achieved by cascading ten such 1-bit stages, and the reconstructed signal is illustrated in Fig.~\ref{fig:nonlinear}.


\subsection{Performance Comparisons}
\label{sec:comp1}

\subsubsection{Comparison with existing NNADCs}We first design an optimal 8-bit NNADC by cascading eight 1-bit stages in Section~\ref{sec:designexp} and compare it with previous NNADCs~\cite{c8,c9}.
The comparative data are summarized in the left columns of Table~\ref{tb:datasets}.
Compared with them, the proposed 8-bit NNADC can achieve the same resolution and higher energy efficiency with ultra-low precision 3-bit RRAM devices.
Both NNADC1 and NNADC2 adopt a typical NN (Hopfield or MLP) architecture to directly train an 8-bit ADC without the optimization of architecture; therefore, they needs high-precision RRAM to achieve the targeted resolution of ADC.
NNADC1 uses a large size ($1\times48\times16$) three-layer MLP as the circuits model, where parasitic aggregations on the large size crossbar array degenerates the conversion speed.
In addition, more hidden neurons are used in NNADC1 which consume more energy.
Since each stage in the proposed 8-bit NNADC resolves only 1-bit and has very small size, it can achieve faster conversion speed with higher energy-efficiency, and high-resolution with low-precision RRAM devices.
Please note that the $FoM_W$ reported in NNADC2 is based on sampling a low frequency (44$\textit{KH}$z) signal at high frequency (1.66$\textit{GH}$z).
Therefore, it is considered outside the scope of a Nyquist ADC, and cannot be compared directly with our work on the same $FoM_W$ basis.
\begin{table*}[!tb]
\footnotesize
\centering
\caption{Performance comparison with different types of ADCs.}
\vspace{-0.2cm}
\begin{threeparttable}
\begin{tabular}{|c|c|c|c|c|c|c|c|c|}
\hline
ADC types         & \multicolumn{3}{c|}{NNADC}            & \multicolumn{3}{c|}{Nonlinear ADC}       & \multicolumn{2}{c|}{Uniform ADC} \\ \hline
Work              & NNADC1\cite{c9}\tnote{*} & NNADC2\cite{c8}\tnote{*}  & \textbf{This work\tnote{*}} & JSSC 09'\cite{c15}\tnote{**}& ISSCC 18'\cite{c3}\tnote{**} & \textbf{This work\tnote{*}} & JSSC 15'\cite{c16}\tnote{**}     & \textbf{This work\tnote{*}}    \\ \hline
Technology ($nm$)   & 130    & 180     & \textbf{130}       & 180      & 90       & \textbf{130}       & 65           & \textbf{130}          \\ \hline
Supply ($V$)  & 1.2    & 1.2     & \textbf{1.5}       & 1.62     & 1.2      & \textbf{1.5}       & 1.2          & \textbf{1.5}          \\ \hline
Area ($mm^2$)         & 0.2    & 0.005$/$0.01 & \textbf{0.02}      & 0.56     & 1.54     & \textbf{0.03}      & 0.594        & \textbf{0.1}          \\ \hline
Power ($mW$)        & 30     & 0.1$/$0.65    & \textbf{25}        & 2.54     & 0.0063   & \textbf{31.3}        & 49.7         & \textbf{67.5}           \\ \hline
$f_S$ ($S/s$)     & 0.3G    & 1.66G$/$0.74G    & \textbf{1G}        & 22M      & 33K      & \textbf{1G}        & 0.25G        & \textbf{1G}           \\ \hline
Resolution (bits) & 8      & 4$/$8       & \textbf{8}         & 8       & 10       & \textbf{10}        & 12           & \textbf{14}           \\ \hline
ENOB (bits)       & 7.96   & 3.7$/$(N$/$A)     & \textbf{8}         & 5.68     & 9.5      & \textbf{9.1}        & 10.6         & \textbf{12.5}           \\ \hline
$FoM_W$ ($fJ/c$)               & 401    & 8.25$/$7.5     & \textbf{97.7}        & 2380     & 263      & \textbf{57}        & 108.5        & \textbf{11.6}           \\ \hline
RRAM precision    & 9      & 6$/$12      & \textbf{3}         & N$/$A      & N$/$A      & \textbf{3}         & N$/$A          & \textbf{3}            \\ \hline
Reconfigurable ?  & Yes    & Yes$/$Yes     & \textbf{Yes}       & No       & Yes      & \textbf{Yes}       & No           & \textbf{Yes}          \\ \hline
\end{tabular}
\begin{tablenotes}\tiny
\item[*] The results are shown based on simulation. 
\item[**] The results are shown on chip.
\end{tablenotes}
\end{threeparttable}
\vspace{-0.7cm}
\label{tb:datasets}
\end{table*}

\subsubsection{Comparison with traditional nonlinear ADCs}We then compare the trained 10-bit logarithmic ADC with state-of-the-art traditional nonlinear ADCs~\cite{c15,c3}.
The comparative data are summarized in the middle columns of Table~\ref{tb:datasets}.
As it shows, the proposed 10-bit logarithmic ADC has competitive advantages in area, sampling rate, and energy efficiency.
JSSC 09'~\cite{c15} uses a pipelined architecture to implement an 8-bit logarithmic ADC.
Due to the devices mismatch, its ENOB degenerates a bit from the targeted resolution.
ISSCC 18'~\cite{c3} requires $>$10-bit capacitive DAC to achieve a configurable 10-bit nonlinear quantization resolution;
therefore, it can achieve high ENOB but only works at $\sim$$\textit{KS}/\textit{s}$ with significant area overhead.
Since we adopt the proposed training framework to directly train a log-encoding signal using small-sized NN models and incorporating device non-idealities, we can achieve a logarithmic ADC with small area, high sampling rate and high ENOB.

\subsubsection{Comparison with traditional uniform ADC}Finally, we compare the trained 14-bit uniform ADC with state-of-the-art traditional uniform ADC.
The comparative data are summarized in the right columns of Table~\ref{tb:datasets}.
It shows that the proposed 14-bit NNADC has competitive advantages in sampling rate, ENOB, and energy efficiency.
JSSC 15'~\cite{c16} uses power hungry op-amps and dedicated calibration techniques, resulting in the power consumption overhead and degeneration of conversion speed.
The proposed 14-bit NNADC uses low-resolution stages with very small NN size, enabling faster conversion speed with higher energy efficiency.
The slight ENOB degeneration of the proposed ADC is caused by the discrepancy (between the trained stage and ideal stage) propagation along the pipeline stages.
Also note that the performance of the proposed NNADCs and the performance of previous NNADCs are based on simulations, while the performance of the traditional nonlinear ADCs and uniform ADC are based on measurements.


\section{Conclusion}

In this paper, we present a co-design methodology that combines a pipelined hardware architecture with a custom NN training framework to achieve high-resolution NN-inspired ADC with low-precision RRAM devices.
A systematic design exploration is performed to search the design space of the sub-ADCs and residue blocks to achieve a balanced trade-off between speed, area, and power consumption of each distinct low-resolution stages.
Using SPICE simulation, we evaluate our design based on various ADC metrics and perform a comprehensive comparison of our work with different types of state-of-the-art ADCs.
The comparison results demonstrate the compelling advantages of the proposed NN-inspired ADC with pipelined architecture in high energy efficiency, high ENOB and fast conversion speed.
This work opens a new avenue to enable future intelligent analog-to-information interfaces for near-sensor analytics using NN-inspired design methodology.

\section*{Acknowledgement}
This work was partially supported by the National Science Foundation (CNS-1657562).


\begin{thebibliography}{99}


\bibitem{c1} R. LiKamWa et al., ``RedEye: Analog ConvNet Image Sensor Architecture for Continuous Mobile Vision," \textit{IEEE ISCA}, 2016, pp. 255-266.


\bibitem{c2} B. Li et al., ``RRAM-Based Analog Approximate Computing," \textit{IEEE TCAD}, vol. 34, no. 12, pp. 1905-1917, 2015.

\bibitem{c3} J. Pena-Ramos et al., ``A Fully Configurable Non-Linear Mixed-Signal Interface for Multi-Sensor Analytics," \textit{IEEE JSSC}, vol. 53, no. 11, pp. 3140-3149, Nov. 2018.

\bibitem{c4} M. Buckler et al., ``Reconfiguring the Imaging Pipeline for Computer Vision," \textit{IEEE ICCV}, 2017, pp. 975-984.






\bibitem{c7} L. Gao et al., ``Digital-to-analog and analog-to-digital conversion with metal oxide memristors for ultra-low power computing," \textit{IEEE/ACM NanoArch}, 2013, pp. 19-22.

\bibitem{c8} L. Danial et al., ``Breaking Through the Speed-Power-Accuracy Tradeoff in ADCs Using a Memristive Neuromorphic Architecture," \textit{IEEE TETCI}, vol. 2, no. 5, pp. 396-409, Oct. 2018.


\bibitem{c9} W. Cao et al., ``NeuADC: Neural Network-Inspired RRAM-Based Synthesizable Analog-to-Digital Conversion with Reconfigurable Quantization Support," \textit{DATE}, 2019, pp. 1456-1461.



\bibitem{c10} T. F. Wu et al., ``14.3 A 43pJ/Cycle Non-Volatile Microcontroller with 4.7$\mu$s Shutdown/Wake-up Integrating 2.3-bit/Cell Resistive RAM and Resilience Techniques," \textit{IEEE ISSCC}, 2019, pp. 226-228.


\bibitem{c11} Y. Cai et al., ``Training low bitwidth convolutional neural network on RRAM," \textit{ASP-DAC}, 2018, pp. 117-122.


\bibitem{c12} H. -. P. Wong et al., ``Metal–Oxide RRAM," \textit{in Proceedings of the IEEE}, vol. 100, no. 6, pp. 1951-1970, June 2012.






\bibitem{c15} J. Lee et al., ``A 2.5mW 80 dB DR 36dB SNDR 22 MS/s Logarithmic Pipeline ADC," \textit{IEEE JSSC}, vol. 44, no. 10, pp. 2755-2765, Oct. 2009.


\bibitem{c16} H. H. Boo et al., ``A 12b 250 MS/s Pipelined ADC With Virtual Ground Reference Buffers," \textit{IEEE JSSC}, vol. 50, no. 12, pp. 2912-2921, 2015.


\bibitem{c17} Kurt Hornik, ``Approximation capabilities of multilayer feedforward networks," \textit{Neural Networks}, vol. 4, issue. 2, pp. 251-257, 1991.


\bibitem{c18} W. Cao et al., ``NeuADC: Neural Network-Inspired Synthesizable Analog-to-Digital Conversion," \textit{IEEE TCAD}, 2019, Early Access.




\bibitem{c20} Kingma et al, ``Adam: A method for stochastic optimization," arXiv preprint arXiv:1412.6980, 2014.






\bibitem{c22} P. Chen and S. Yu, ``Compact Modeling of RRAM Devices and Its Applications in 1T1R and 1S1R Array Design," \textit{IEEE TED}, vol. 62, no. 12, pp. 4022-4028, Dec. 2015.



\bibitem{c23} B. Li, et al., ``MErging the Interface: Power, area and accuracy co-optimization for RRAM crossbar-based mixed-signal computing system," \textit{IEEE ACM/EDAA/IEEE DAC}, 2015, pp. 1-6.


\bibitem{c25} Weidong Cao et al., ``A 40Gb/s 39mW 3-tap adaptive closed-loop decision feedback equalizer in 65nm CMOS," \textit{IEEE MWSCAS}, 2015, pp. 1-4.



\end{thebibliography}
\end{document}